%% file: My_Main-1.tex
\documentclass[sigconf]{acmart}

\AtBeginDocument{%
  }

\usepackage{multirow}
\usepackage{algorithm}
\usepackage{algorithmic}

\usepackage{amsmath}
\usepackage{mathtools} 
\usepackage{booktabs} 

\usepackage{makecell}
\usepackage{adjustbox} 
\usepackage{colortbl}
\usepackage{xcolor}

\begin{document}

\title{Groc-PO: GROunded Context Preference Optimization for Truthful Multimodal LLMs}

\author{Zhixiao Zheng}
\email{zhixiao.zheng@mail.ustc.edu.cn}
\affiliation{
  \institution{University of Science and Technology of China}
    \city{Hefei}
  \country{China}
}

\author{Zheren Fu}
\email{fzr@ustc.edu.cn}
\authornote{Corresponding author.}
\affiliation{
  \institution{University of Science and Technology of China}
    \city{Hefei}
  \country{China}
}

\author{Zhiyuan Yao}
\email{yaozhiyuan@mail.ustc.edu.cn}
\affiliation{
  \institution{University of Science and Technology of China}
  \city{Hefei}
  \country{China}
}

\author{Chunxiao Liu}
\email{chunxiao6liu@gmail.com}
\affiliation{
  \institution{Xiaomi Corporation}
  \city{Beijing}
  \country{China}
}

\author{Dongming Zhang}
\email{zhangdongming@people.cn}
\affiliation{
  \institution{State Key Laboratory of Communication Content Cognition, People's Daily Online}
    \city{Beijing}
  \country{China}
}

\author{Zhendong Mao}
\email{zdmao@ustc.edu.cn}
\affiliation{
  \institution{University of Science and Technology of China}
    \city{Hefei}
  \country{China}
}

\begin{abstract}
Despite the rapid progress of Multimodal Large Language Models (MLLMs), they still suffer from untruthfulness issues, such as visual hallucinations, content fabrication, and unfaithful reasoning, which substantially undermine their faithfulness and practical utility. Alignment methods based on human preference, such as Direct Preference Optimization (DPO), have been widely adopted to address these issues. 
However, multimodal reasoning errors often propagate across stages, and final-answer errors can often be traced to mistakes in early grounding stages, yet standard DPO typically applies preference optimization at the final-answer level. 
This credit-assignment challenge means that supervision for early grounding stages is indirect rather than stage-specific, making it difficult to suppress error propagation arising from grounding drift and context inconsistency. 
To address this, we propose Grounded Context Preference Optimization (Groc-PO), a grounded preference optimization framework for MLLMs. We further construct the Grounded Context Preference Dataset (GCPD), organizing multi-stage preference samples around three stages of Object Grounding, Contextual Grounding, and Grounded Reasoning, to capture the formation, integration, and utilization of grounded context. 
By introducing more explicit preference supervision over multiple grounded stages, Groc-PO strengthens context-dependent reasoning and mitigates cross-stage error propagation. Extensive experiments show that, compared with standard DPO and other strong baselines, Groc-PO achieves improved performance in hallucination mitigation, faithful reasoning, and overall reliability, supporting the value of more explicit grounded supervision for trustworthy multimodal reasoning.
\end{abstract}

\ccsdesc[500]{Computing methodologies~Machine learning}

\keywords{Multimodal LLMs, Preference Optimization, Truthful Models}


\maketitle

\input{sec/introduction}
\input{sec/related}

\input{sec/model}

\input{sec/experiment}

\section{Conclusion}
In this paper, we proposed Grounded Context Preference Optimization (Groc-PO), a framework that improves MLLM faithfulness through explicit preference supervision over grounded pre-final stages. To support this objective, we introduced the Grounded Context Preference Dataset (GCPD) and novel adaptive loss function. Extensive evaluations demonstrate that Groc-PO comprehensively enhances multiple capabilities, contributing to developing more faithful MLLMs.

\section{Acknowledgements}
This work was supported by the Artificial Intelligence-National Science and Technology Major
Project (2023ZD0121200) and the Fundamental and Interdisciplinary Disciplines Breakthrough Plan of
the Ministry of Education of China (No. JYB2025XDXM103).

\bibliographystyle{ACM-Reference-Format}
\bibliography{My_sample-base}
\end{document}

%% file: sec/introduction.tex
\section{Introduction}
\label{sec:intro}

\begin{figure}[t!]
  \centering
  \includegraphics[width=\linewidth]{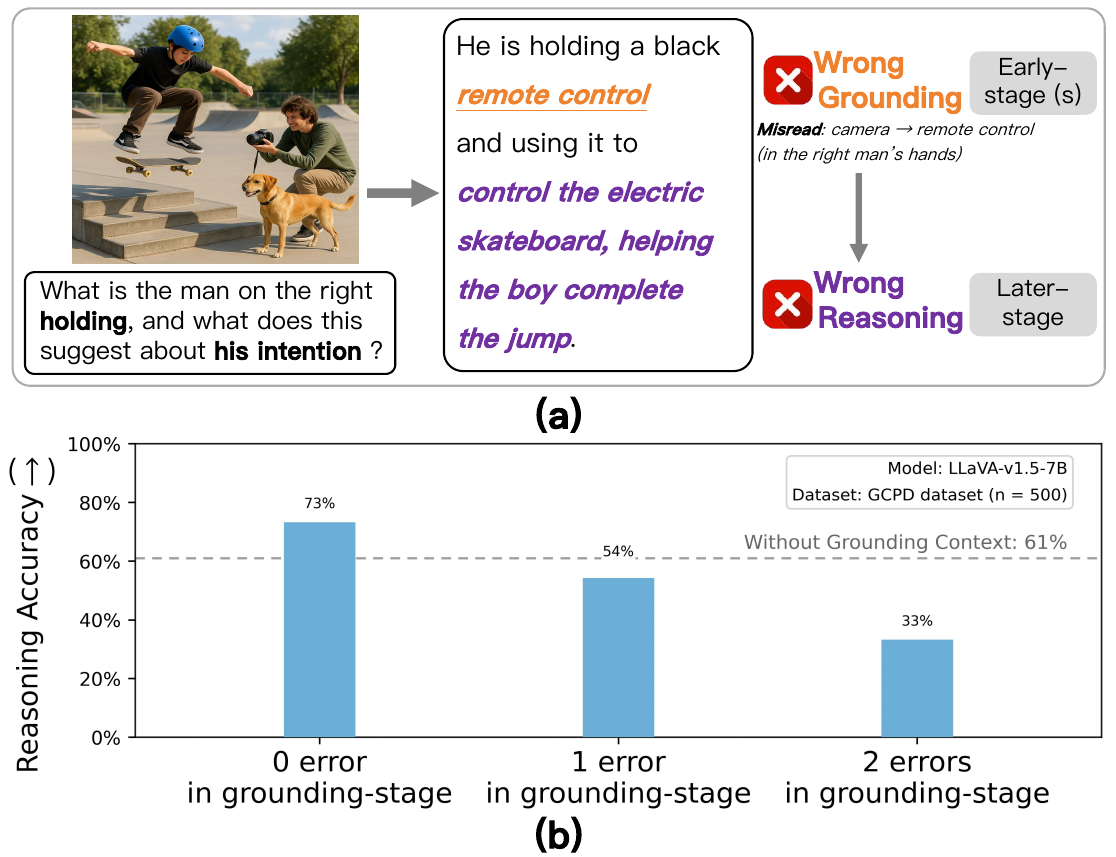}
  \caption{Motivating example of error propagation across stages in MLLMs. (a) A case where an early grounding error propagates to the later reasoning stage and leads to an incorrect answer. (b) Statistical experiments with LLaVA-v1.5-7B~\cite{liu2023llava} on GCPD dataset (constructed from RLHF-V~\cite{yu2024rlhf}), showing that introducing errors into 0, 1, or 2 grounding stages is associated with progressively lower final reasoning accuracy, consistent with error propagation in MLLMs.}
  \label{fig:motivation_1}
\end{figure}

Multimodal Large Language Models (MLLMs), by integrating powerful vision and language capabilities, are transforming human-computer interaction~\cite{alayrac2022flamingo, liu2024improved, amirloo2024understanding}. MLLMs also demonstrate astonishing potential in tasks like Visual Question Answering (VQA) and medical analysis~\cite{xiao2024detecting, pi2024strengthening, zhang2024automated}. However, despite these advancements, MLLMs suffer from the unfaithfulness problems, such as generating content that contradicts visual facts (e.g., fabricated objects, attributes, or relationships)~\cite{bai2024hallucination, gunjal2024detecting}. This significantly hinders their reliability and utility in real-world applications~\cite{liu2024survey, liang2024survey}.

To mitigate the above issues, preference learning based on human feedback has become a mainstream alignment paradigm. Among these methods, Direct Preference Optimization (DPO)~\cite{rafailov2023direct} has been widely adopted because it directly optimizes the policy from preference data without requiring an explicit reward model. Existing DPO methods apply preference supervision at the final-answer-level to guide the model in generating more accurate and reliable outputs. 
However, multimodal reasoning often involves several stages, including early grounding and later reasoning, and the quality of the final answer depends on the robustness of the full process.
As illustrated in Fig. ~\ref{fig:motivation_1}a, 
an error arising in an early grounding stage can propagate to later reasoning stages and lead to reasoning failure.
Therefore, it remains unclear whether standard DPO, which mainly relies on holistic final-answer-level preferences, can adequately address reasoning errors rooted in early grounding stages.

To this end, we conduct further controlled explorations. As shown in Fig.~\ref{fig:motivation_1}b, introducing errors into more grounding stages is associated with lower final reasoning accuracy, indicating that errors can propagate and accumulate across stages along the reasoning process. This observation reveals a key structural limitation of final-answer-level preference optimization, namely the propagation of MLLM errors: later reasoning failures often do not arise solely at the answer-generation stage, but instead originate from early grounding stages and then accumulate along the reasoning process, affecting the final reasoning outcome. Yet standard DPO typically supervises only at the final-answer-level, offering holistic guidance while providing little stage-specific supervision for these early grounding stages.

Consequently, an important question for MLLMs preference alignment is how to provide more targeted preference supervision for early grounding stages, so as to reduce error propagation and improve faithful multimodal reasoning.
To address these problems, we propose Grounded Context Preference Optimization (Groc-PO), a grounded preference optimization framework for MLLMs. 
Built upon the newly constructed Grounded Context Preference Dataset (GCPD), Groc-PO leverages preference signals from three contextual stages: Object Grounding, Contextual Grounding, and Grounded Reasoning. This design provides more explicit supervision for upstream grounded stages, thereby enhancing the model's ability to robustly construct and faithfully utilize the grounded context. 
The framework incorporates the full context and adopts an adaptive stage-aware optimization strategy, improving complex reasoning while mitigating the propagation of upstream errors. Extensive experiments demonstrate that Groc-PO yields notable gains in hallucination mitigation, contextual understanding, and faithful reasoning. The main contributions of this paper are as follows:
\begin{itemize}
    \item We propose the Groc-PO framework. By introducing explicit supervision over grounded pre-final stages, it improves faithful multimodal reasoning under multi-round contexts and effectively mitigates error propagation in MLLMs.
    \item We construct the Grounded Context Preference Dataset (GCPD). GCPD organizes multi-stage preference data around Visual Grounding, Context Grounding, and Faithful Complex Reasoning, providing support for explicit stage-wise supervision of grounded context.
    \item Built on GCPD, Groc-PO employs an adaptive grounded preference optimization mechanism that dynamically allocates learning emphasis across different stages and sample complexities, enabling more targeted alignment.
    \item We conduct systematic experiments across multiple datasets and benchmarks. The results show that Groc-PO consistently outperforms standard DPO and several strong baselines on hallucination and complex capability evaluations, validating the effectiveness of explicit supervision for pre-final grounded stages.
\end{itemize}

%% file: sec/related.tex
\section{Related Works}
\label{sec:related_works}

\begin{figure*}[t!]
  \centering
  \includegraphics[width=1.0\linewidth]{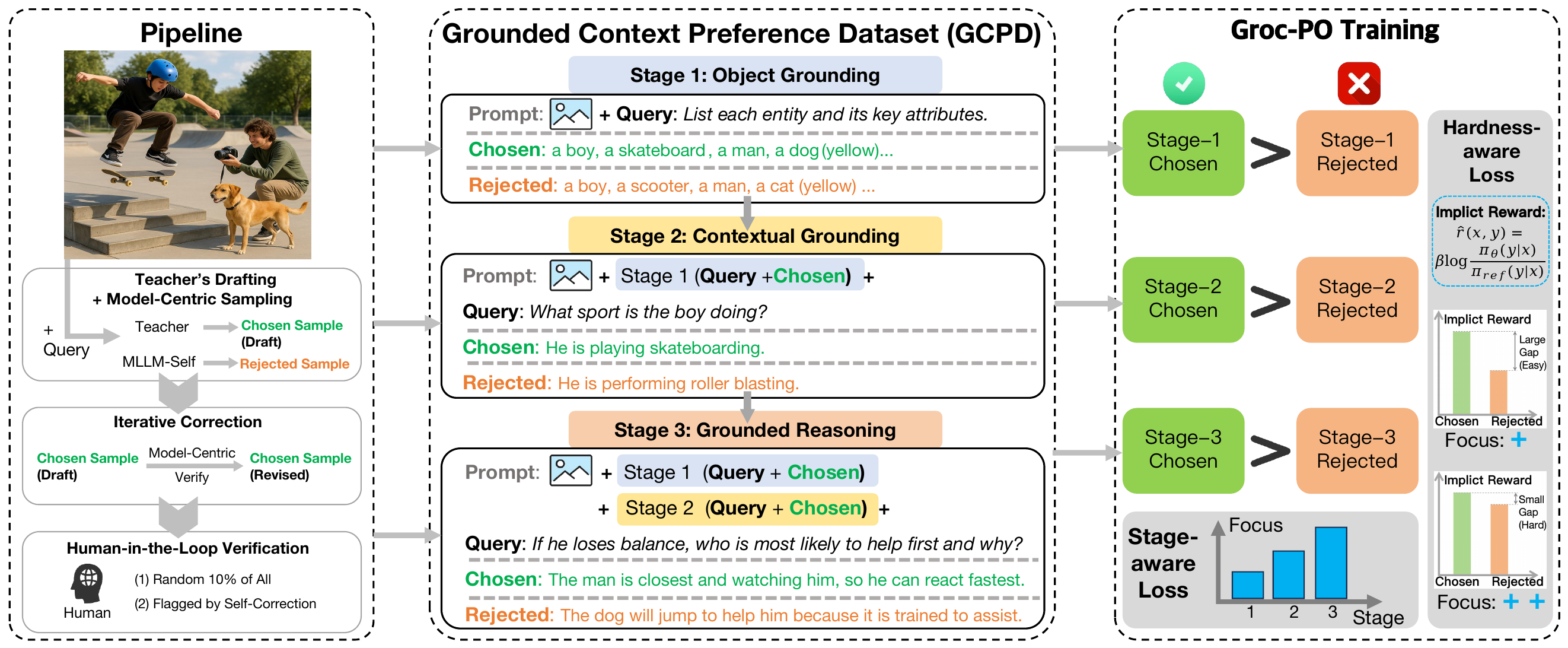}
    \caption{Overview of our Grounded Context Preference Optimization (Groc-PO) framework, including GCPD dataset construction. The left panel shows dataset construction, where multi-stage preference pairs are generated through teacher-assisted drafting, model-centric sampling, iterative correction, and human verification. The middle panel presents three stages of grounded preference supervision: Stage 1 for object grounding, Stage 2 for contextual grounding, and Stage 3 for grounded reasoning. The right panel shows Groc-PO training, which jointly uses preference pairs from all three stages with a stage-aware, hardness-aware loss to improve context-dependent reasoning and mitigate cross-stage error propagation.}
  \label{fig:model}
\end{figure*}

\subsection{Unfaithfulness in MLLMs}
Unfaithfulness (or hallucination) in MLLMs refers to the generation of content inconsistent with the visual input, typically manifested as fabricated objects, incorrect attributes, or misinterpreted relationships~\cite{bai2024hallucination}.
Hallucinations stem from training data flaws~\cite{li2023evaluating}; module biases~\cite{guan2024hallusionbench}; suboptimal training paradigms~\cite{ben2023mocha}; and inference‑stage defects~\cite{huang2024opera}.
To address hallucinations, approaches fall into two categories: training‑free methods (e.g., Opera~\cite{huang2024opera}, VCD~\cite{leng2024mitigating}) and training‑based techniques (e.g., RLHF, PPO)~\cite{xiao2025detecting,schulman2017proximal}.

\subsection{Preference Learning for faithful MLLMs}

Preference learning initially applied to LLM alignment via RLHF, but DPO has recently gained widespread adoption as a simpler and stable alternative.
V-DPO~\cite{xie2024v} extends DPO by incorporating visual context learning. POVID~\cite{zhou2024aligning} creates a fine-grained dataset by injecting noise to texts and images. RLHF-V~\cite{yu2024rlhf} collects segment-level human preference data and performs dense DPO. 
In addition, SPO-Task Planning~\cite{liang2025structured}
constructs preference pairs using curriculum learning to improve long-horizon planning.
MM-RLHF~\cite{zhang2025mm} constructed a preference dataset and proposed a novel reward model to achieve MLLM alignment.
SPO~\cite{sun2025structured} treats questioning and answering jointly as a policy trajectory, co-optimizing them via a structured reward function to enhance the model's consideration of visual dependency in dialogue. The mrDPO~\cite{tang2024enhancing} utilizes multi-round DPO and Rebirth Tuning to optimize audio-visual LLMs.

%% file: sec/model.tex
\section{Methodology}
\label{sec:model}

The overview of our Groc-PO framework is illustrated in Fig. \ref{fig:model}. We first introduce the DPO, followed by the novel Grounded Context Preference Dataset (GCPD) and its generation pipeline, and the adaptive Groc-PO Loss.

\subsection{Preliminaries: Direct Preference Optimization}
\label{subsec:dpo_preliminaries}
DPO directly optimizes the model through a contrastive learning objective, making it more inclined to generate human-preferred responses while reducing the probability of generating dispreferred responses. 
DPO learns from preference data $(x, y^{+}, y^{-}) \sim \mathcal{D}$, where $x$ is the input prompt, $y^{+}$ is the human-preferred /chosen response, $y^{-}$ is the dispreferred /rejected response, and $\mathcal{D}$ is the dataset.

The DPO objective function
assumes that the human preference probability $p^*(y^{+} \succ y^{-} \mid x)$ can be modeled via a latent reward function $r^*(x,y)$: $p^*(y^{+} \succ y^{-} \mid x) = \sigma(r^*(x,y^{+}) - r^*(x,y^{-}))$.
DPO further relates the reward function to the model's policy $\pi_{\theta}$ and a reference policy $\pi_{\text{ref}}$: 
$r_{^*}(x,y) = \beta (\log(\pi_{\theta}(y \mid x)) - \log(\pi_{\text{ref}}(y \mid x)))$.

where $\beta$ is a hyperparameter controlling the ratio between reward function and policy deviation. DPO's loss can directly optimize MLLM to maximize the probability of generating $y^{+}$ and minimize generating $y^{-}$. 
Let us define the log-likelihood ratio for the preferred response as $r^{+} = \log\left( \pi_\theta(y^{+}\mid x) / \pi_{\mathrm{ref}}(y^{+}\mid x) \right)$ and for the dispreferred response as $r^{-} = \log\left( \pi_\theta(y^{-}\mid x) / \pi_{\mathrm{ref}}(y^{-}\mid x) \right)$. Then the DPO loss function is defined as:

\begin{equation}
\label{eq:dpo}
\mathcal{L}_{\mathrm{DPO}} = -\log\sigma\!\left(\beta(r^+ - r^-)\right).
\end{equation}

By minimizing this loss function, the model $\pi_{\theta}$ is trained to increase the difference between the log-probabilities of $y^{+}$ and $y^{-}$,
It makes DPO simpler and demonstrates comparable or superior performance to RLHF.
\subsection{Grounded Context Preference Dataset (GCPD)}
\label{sec:gcpd_dataset}
Failures in multimodal reasoning often do not emerge only at the final response, but can originate from imperfect grounding and accumulated inconsistencies in pre-final stages. To explicitly supervise these upstream stages, we construct the \textbf{Grounded Context Preference Dataset (GCPD)}, a structured preference dataset organized around progressively accumulated grounded context.

Specifically, GCPD is built as a 3-stage context-dependent preference dataset. Its three stages move from basic visual grounding, to context-grounded understanding, and finally to faithful complex reasoning. This formulation places preference supervision not only on the final answer, but also on the grounded context that supports it, providing a more direct signal for reducing error propagation and improving multimodal faithfulness.

\textbf{Cumulative Multi-stage Context:} For any stage $s$, the prompt includes all historical context from stage 1 to $s-1$, ensuring continuous information flow. Let $\mathit{I}$ denote the image, $\mathit{Q}$ denote the question, $\mathit{A}$ denote the answer, $\mathit{A}_{s}^+$ means the chosen response (human-preferred). Then, we will have the following prompt structure for every stage:
\begin{itemize}
    \item \textbf{Stage 1 (S1)}: Prompt $= \{I, Q_1\}$
    \item \textbf{Stage 2 (S2)}: Prompt $= \{I, Q_1, A_{s_1}^+, Q_2\}$
    \item \textbf{Stage 3 (S3)}: Prompt $= \{I, Q_1, A_{s_1}^+, Q_2, A_{s_2}^+, Q_3\}$
\end{itemize}

Then, we detail the progressive stages design:
\begin{itemize}
    \item \textbf{S1: Object Grounding.} S1 marks the starting point of CoT process—identifying basic facts. We present the model with a standardized question (e.g., List each entity and its key attributes) to extract core visual elements.
    
    \item \textbf{S2: Contextual Grounding.} S2 simulates the intermediate steps of CoT. 
    Building upon S1, S2 focuses on tasks such as relationship description, comprehensive captioning, or visual question answering. It requires the model not only to identify individual entities but also to understand how they form a meaningful whole.
    
    \item \textbf{S3: Grounded Reasoning.} S3 is the culmination of the CoT simulation. We pose complex questions that require integrating the image with context from S1 and S2, and performing logical inference, intent prediction, or reasoning tasks. This compels the model to perform high-level cognition based on established, reliable context.
\end{itemize}
In this way, our GCPD dataset is no longer fragmented question-answer pairs but offers a progressive learning process aligned with human cognitive laws.

\subsection{Pipeline of GCPD's Generation and Features}
\label{sec:data_pipeline}
\subsubsection{Basic Workflow}
Our pipeline begins with a widely recognized RLHF-V dataset~\cite{yu2024rlhf}, which contains 5,733 images along with human-annotated, high-quality preference pair for tasks like captioning, relational description. Our goal is to generate a structured 3-stage context for each image. For any stage $s$, the core tasks is to generate a specific question $Q_s$ and a high-quality preference pair $(y_s^+, y_s^-)$, consisting of a chosen and a rejected response.
This three-stage structure is designed with progressively increasing complexity, following a perception-understanding-reasoning path.

For the \textbf{Stage 1}, we define a universal base question $\mathbf{Q}_1$: ``List each entity and its key attributes,'' to establish a factual foundation. The chosen response, $y_{s_1}^+$, is initially generated by an advanced teacher model; the rejected response, $y_{s_1}^-$, is derived from $y_{s_1}^+$ via introduced rule-based deficiencies, retaining structural soundness while including factual inaccuracies.
Subsequently, $y_{s_1}^+$ undergoes rigorous verification and refinement to ensure its accuracy.

For the \textbf{Stage 2}, aiming for both efficiency and quality, we adopt the existing data from the RLHF-V dataset corresponding to each image, as its design philosophy aligns perfectly with our goals for this stage. This means that the question $Q_2$, the chosen response $y_{s_2}^+$, and the rejected response $y_{s_2}^-$ are all sourced from this high-quality, human-validated dataset. It guarantees the superiority of the data for relational description and understanding.

In the \textbf{Stage 3}, our objective is to enhance model's ability of complex thinking. We leverage an advanced teacher model to generate a new, more profound, and complex question, $Q_3$, based on the context of the original image and the preceding dialogue ($Q_1, y_{s_1}^+, Q_2, y_{s_2}^+$). Following the pattern of the 1st stage, we then generate a high-quality chosen response, $y_{s_3}^+$, via the teacher model combined with rigorous verification and refinement, and a rejected response, $y_{s_3}^-$, from our LLaVA base model to complete the final preference pair.

\subsubsection{Iterative Self-Correction for Chosen Samples ($y_s^+$)}
To maximize the chosen response quality and factual accuracy, we introduce an iterative self-correction mechanism.
\begin{itemize}
    \item In Stage 1, the initial list of entities in the chosen response, generated by the advanced teacher model, is fed back into the teacher model with a detailed verification prompt. The teacher model is instructed to comprehensively check and provide a score. If the response contains hallucinations or the average score is too low, it triggers a rewrite by the advanced teacher model.
    \item Similarly, in Stage 3, the complex reasoning answer in the chosen response is sent back for a second review. In this step, the teacher model acts as a \texttt{"}Critic\texttt{"}, inspecting the reasoning chain for logical fallacies and ensuring it is fully grounded in the provided visual and textual context.
\end{itemize}
This \texttt{"}generate-and-refine\texttt{"} closed-loop process significantly enhances the quality of our chosen responses, providing the model with a clear and reliable learning target.

\subsubsection{Model-Centric Sampling}
\label{ssec:model_centric_sampling}
To further improve data quality and make training more efficient, we draw inspiration from the \textit{on-policy} concept in RLHF~\cite{christiano2017deep} and adopt a \textit{Model-Centric Sampling} strategy. The core idea is to ensure that the distribution of training data, 
aligns as closely as possible with the generation distribution of our targeted fine-tuned model. 
This approach enables the model to directly confront and rectify its own predominant error patterns, making the fine-tuning process highly targeted.
This strategy is reflected in two key aspects:
\begin{itemize}
    \item The rejected responses in Stage 3 ($y_{s_3}^-$) are generated by our target LLaVA model. Consequently, these samples are representative of the model's intrinsic failure modes, particularly in areas like long-range dependency, contextual understanding, and complex reasoning, which manifest as logical fallacies or cumulative hallucinations. In contrast to negative samples from a more capable, external model teacher model, which often suffer from a distribution mismatch in $y_r^-$. These 'model-centric' samples provide a highly targeted and valuable learning signal for DPO.
    \item In Stage 2, a key characteristic of the RLHF-V dataset lies in its preference pair construction: the rejected responses ($y_{s_2}^-$) are generated by the MLLM family, while the chosen responses ($y_{s_2}^+$) are human-revised versions of these same rejected samples. 
    This approach ensures high distributional and stylistic alignment with our target model, effectively forming a tailored \texttt{"}problem-solution\texttt{"} paradigm for its specific weaknesses. This mechanism provides the highly valuable and targeted learning signal that is the core rationale for our adoption of this dataset.

\end{itemize}

It is worth noting that the Stage 1 responses are highly uniform, making generation variance across models minimal, and thus Model-Centric Sampling has little impact.

\subsubsection{Human-in-the-Loop Verification}
To ensure the rigor and quality of our GCPD dataset, we introduced Human-in-the-Loop verification. The audit team consisted of three MLLM-familiar PhD students who adhered to a guideline for all checks and corrections.
The samples reviewed included: first, a 10\% random sample of the entire dataset; second, a targeted review of $y_s^+$ samples flagged by the teacher model Critic as having major issues during the \texttt{"}Iterative Self-Correction\texttt{"} process.

Overall, approximately 12\% ($\sim$2k) of the $y_s^+$ samples were manually audited, leading to the revision or rewriting of nearly 2\% of severely problematic samples (primarily in R3). The total time cost was approximately 57 hours per reviewer (30h for auditing, 27h for revision). This mechanism ensures the reliability and faithfulness of the dataset.

\newcolumntype{C}[1]{>{\centering\arraybackslash}m{#1}}
\begin{table*}[t]
\small
\caption{\textbf{Performance comparison with leading methods on various hallucination and general benchmarks.} On LLaVA-v1.5-7B, LLaVA-v1.5-13B~\cite{liu2023llava}, Groc-PO achieves significant leads on key faithfulness metrics (e.g., AMBER, MM-Hal) and simultaneously enhances general abilities (e.g., LLAVA-Bench, SEED). Here, AMBER-Gene. refers to AMBER-Generation, and AMBER-Discri. denotes the AMBER-Discrimination.}
\label{tab:main_results}
\centering
\setlength{\tabcolsep}{4pt} 
\renewcommand{\arraystretch}{1.1}
\begin{tabular}{@{}l cc ccc cc ccc@{}} %
\toprule
\multirow{2}{*}{\makecell{\textbf{Methods}}} & \multicolumn{2}{c}{\textbf{MM-Hal}~\cite{sun2023aligning}} & \multicolumn{3}{c}{\textbf{AMBER-Gene.}~\cite{wang2023llm}} & \multicolumn{2}{c}{\textbf{AMBER-Discri.}~\cite{wang2023llm}} & \multirow{2}{*}{\makecell{\textbf{LLaVA}~\cite{liu2023visual}}} & \multirow{2}{*}{\makecell{\textbf{SEED}~\cite{li2023seed}}} \\
\cmidrule(lr){2-3} \cmidrule(lr){4-6} \cmidrule(lr){7-8}
& Score$\uparrow$ & Hal-Rate$\downarrow$ & CHAIR$\downarrow$ & Hal-Rate$\downarrow$ & Cog$\downarrow$ & Acc$\uparrow$ & F1$\uparrow$ & & & \\ 
\midrule
\textbf{LLaVA-1.5-7B}~\cite{liu2023llava} & 2.01 & 61.4 & 7.8 & 36.4 & 4.2 & 71.7 & 74.3 & 65.6 & 66.1 \\
+ DPO \citep{rafailov2023direct} & 2.14 & 58.3 & 5.7 & 27.3 & 2.6 & 71.3 & 82.1 & 69.1 & 66.4 \\
+ CSR \citep{zhou2024calibrated} & 2.05 & 60.4 & 5.4 & 25.5 & 2.6 & 73.2 & 76.1 & 68.9 & 65.9 \\
+ POVID \citep{zhou2024aligning} & 2.26 & 55.2 & 5.7 & 26.9 & 3.0 & 71.9 & 74.7 & 68.2 & 66.1 \\
+ V-DPO \citep{xie2024v} & 2.16 & 56.0 & 5.6 & 27.3 & 2.7 & - & 81.6 & - & - \\
+ RLHF-V \citep{yu2024rlhf} & 2.02 & 60.4 & 5.5 & 26.3 & 2.5 & 74.8 & 78.5 & 68.0 & 66.1 \\
+ mDPO~\citep{wang2024mdpo} & 2.39 & 54.0 & 4.4 & \textbf{24.5} & 2.4 & - & - & - & - \\
\rowcolor{cyan!12}
+ \textbf{Groc-PO} (Ours) & \textbf{2.76} & \textbf{47.0} & \textbf{4.2} & 25.2 & \textbf{1.5} & \textbf{78.0} & \textbf{85.0} & \textbf{72.9} & \textbf{67.1} \\
\midrule 
\textbf{LLaVA-1.5-13B}~\cite{liu2023llava} & 2.38 & 53.1 & 7.0 & 33.1 & 3.3 & 71.4 & 73.1 & 73.1 & 68.2 \\
+ DPO \citep{rafailov2023direct} & 2.47 & 51.0 & 6.1 & 26.3 & 2.7 & 71.9 & 82.1 & 72.8 & 68.6 \\
+ RLHF-V \citep{yu2024rlhf} & 2.50 & 52.1 & 6.3 & 25.1 & 2.1 & 79.2 & 82.3 & \textbf{76.7} & 68.2 \\
\rowcolor{cyan!12}
+ \textbf{Groc-PO} (Ours) & \textbf{2.85} & \textbf{45.0} & \textbf{3.8} & \textbf{24.4} & \textbf{1.3} & \textbf{83.5} & \textbf{88.2} & 76.5 & \textbf{68.8}  \\
\bottomrule
\end{tabular}
\end{table*}

\subsection{Customized Groc-PO Loss}
\label{sec:grocpo_loss}
The standard DPO loss (Equation \ref{eq:dpo}) treats all samples in the dataset equally. This uniform approach overlooks the inherent gradient of cognitive depth and sample difficulty within our GCPD dataset.
To better leverage this rich and structured information, we propose a Groc-PO Loss, with a sample-level adaptive weight, $w_i$, enabling the model to dynamically focus on samples that are more informative and have higher learning value. The Groc-PO loss is defined as:
\begin{equation}
\mathcal{L}_{\text{Groc-PO}} = -\mathbb{E}_{(x_i, y_{w,i}, y_{l,i}) \sim \mathcal{D}} \left[ w_i \cdot \log\sigma(r_i(\theta)) \right]
\end{equation}
where $r_i(\theta) = \beta \log \frac{\pi_{\theta}(y_i^+|x_i)}{\pi_{\text{ref}}(y_i^+|x_i)} - \beta \log \frac{\pi_{\theta}(y_i^-|x_i)}{\pi_{\text{ref}}(y_i^-|x_i)}$ is the implicit reward difference for preference pair $i$. The core innovation lies in the design of adaptive weight $w_i$, which is composed of two multiplicative components:
\begin{equation}
w_i = \lambda_{r(i)} \cdot \gamma_i
\end{equation}

\subsubsection{Stage-aware Importance Weight ($\lambda_r$)}
This weight is designed to reflect the learning value of different rounds.
Later rounds represent more complex tasks that demand stronger long-range dependency and comprehensive abilities. Thus, for any stage $s$, we design $\lambda_s$ as a monotonically increasing function of $s$ to encourage model to 
focus on these advanced knowledge: 
\begin{equation}
\lambda_r = 1 + \alpha(r-1)
\end{equation}
where $r \in \{1, 2, 3\}$ is the dataset stage, and $\alpha \ge 0$ is a hyperparameter that controls the growth rate of stage importance. When $\alpha > 0$, samples from the latter 2 stages are assigned a higher loss weight.

\subsubsection{Hardness-aware Focusing Weight ($\gamma_i$)}
This weight 
aims to make model focus more on \texttt{"}hard samples\texttt{"} that are difficult to distinguish. When the model can easily distinguish between $y^+$ and $y^-$ , the sample is \texttt{"}easy\texttt{"} and has low learning value. Conversely, when model perceives two responses as having similar quality, the sample is \texttt{"}hard\texttt{"} and should be prioritized. We define $\gamma_i$ as: 
\begin{equation}
\gamma_i = (1 - \sigma(r_i(\theta)))^{\eta}
\end{equation}
where $\eta \ge 0$ is a focusing parameter. For $\eta > 0$, this term significantly decreases the loss for well-distinguished samples (where $\sigma(r_i) \to 1$), thereby directing the optimization process toward the most challenging pairs.

Through this dual-weighting mechanism, our Groc-PO Loss adaptively evaluates the importance of each training sample, considering both its role in the progressive rounds (via $\lambda_t$) and sample's learning value (via $\gamma_i$).

%% file: sec/experiment.tex
\section{Experiments}

\subsection{Datasets, Metrics and Implementation Details}
\label{sec:datasets_metrics}
\paragraph{\textbf{Training Data:}}
Based on our GCPD dataset generation pipeline described in Section 3.2, we constructed the 3-stages preference dataset comprising 5,733 diverse images and 17,199 high-quality preference pairs.
\paragraph{\textbf{Evaluation Benchmarks:}}
To comprehensively evaluate models' performance,
we employ several widely used benchmarks:
\textbf{For faithfulness evaluation}, 
\textbf{AMBER}~\cite{wang2023llm} is a LLM-free benchmark for evaluating hallucinations, which has two components: 
(a) \textit{Discrimination:} deciding whether a statement is correct; (b) \textit{Generation:} describing for an image.
\textbf{MM-Hal}~\cite{sun2023aligning} evaluates response-level hallucination rate and informativeness.
\textbf{For general capability evaluation},
\textbf{LLaVA-Bench}~\cite{liu2023visual} is a benchmark spanning diverse scenarios.
\textbf{SEED-Bench}~\cite{li2023seed} is a large-scale benchmark to assess model abilities likes visual understanding and reasoning.
\textbf{For complex tasks}, some sub-tasks of benchmarks are adopted, such as \textit{LLaVA-Bench-complex reasoning}~\cite{liu2023visual}, \textit{LLaVA-Bench-conversation}~\cite{liu2023visual}, \textit{SEED-visual reasoning}~\cite{li2023seed}, \textit{MME-commonsense-reasoning}~\cite{fu2023mme}, and multi-turn dialogue benchmark \textit{MM-MT}~\cite{agrawal2024pixtral}.

\paragraph{\textbf{Implementation Details:}}
\label{sec:implementation_details}
Our experiments leverage the widely adopted LLaVA-v1.5-7B and 13B~\cite{liu2023llava} and Qwen2.5-VL-7B~\cite{Qwen2.5-VL} models to evaluate scalability and effectiveness of our method. We employed LoRA~\cite{hu2022lora} and AdamW optimizer~\cite{loshchilov2017decoupled}. Training was performed over 2 epochs with an effective batch size of 32. The teacher model we used is GPT-4o~\cite{hurst2024gpt}.

\subsection{Main Results}
\label{sec:main_results}
On multiple mainstream faithfulness and general benchmarks, we conduct a comprehensive comparison of Groc-PO with a series of representative baselines, such as llava-v1.5-7B~\cite{liu2023llava}, DPO~\cite{rafailov2023direct}, CSR~\cite{zhou2024calibrated}, POVID~\cite{zhou2024aligning}, RLHF-V~\cite{yu2024rlhf}, V-DPO~\cite{xie2024v}, and mDPO~\citep{wang2024mdpo}.

\paragraph{\textbf{On Faithfulness and General Abilities Evaluation.}}
Table \ref{tab:main_results} shows that Groc-PO achieves leading performance across almost all key evaluation metrics. These evaluations cover different types of faithfulnesss and general ability tests. These demonstrate that through our progressive preference data and adaptive training framework, the model can not only significantly suppress hallucinations but also improve comprehensive abilities

Furthermore, Groc-PO shows its scalability by delivering consistent and substantial gains across models from 7B to 13B, validating its efficiency and broad applicability.

In addition, to validate the scalability and generalizability of our framework, we applied Groc-PO to Qwen2.5VL-7B~\cite{Qwen2.5-VL}. As detailed in Table \ref{tab:Qwen2.5vl-7b}, the model achieved improvements on the faithfulness test.

\paragraph{\textbf{On Complex Understanding and Reasoning.}}
Table \ref{tab:reason} demonstrates the particularly prominent superiority of Groc-PO in complex reasoning tasks. Across demanding benchmarks, including LLaVA-bench-complex, LLaVA-conversation, SEED-reasoning, and MME-Commonsense-reasoning, our model consistently outperforms the baseline, achieving up to a 45\% relative improvement.

Furthermore, Groc-PO demonstrates superior conversational capabilities on MM-MT benchmark (multi-round dialogue), validating the effectiveness of its structured context.

\vspace{-0.5em}
\begin{table}[t]
\centering
\small
\caption{Comparison with different preference optimization methods on Qwen2.5-VL-7B~\cite{Qwen2.5-VL}.}
\label{tab:Qwen2.5vl-7b}
\vspace{-0.8em}
\begin{tabular}{lcc}
\toprule
Method & MM-Hal Score ($\uparrow$) & MM-Hal Rate ($\downarrow$) \\
\midrule
Qwen2.5-VL-7B~\cite{Qwen2.5-VL} & 3.55 & 0.40 \\
\quad + DPO~\cite{rafailov2023direct} & 3.59 & 0.38 \\
\quad + POVID~\cite{zhou2024aligning}  & 3.73 & 0.37 \\
\quad + CSR~\cite{zhou2024calibrated}  & 3.71 & 0.41 \\
\rowcolor{cyan!12}
\quad + \textbf{Groc-PO (Ours)} & \textbf{3.83} & \textbf{0.32} \\
\bottomrule
\end{tabular}
\vspace{-0.5em}
\end{table}

\begin{table}[t]
  \caption{\textbf{Comparison of Groc-PO and DPO on complex tasks}. Tasks cover reasoning (\textit{LLaVA-Bench-complex reasoning}, \textit{SEED-visual reasoning}, \textit{MME-commonsense-reasoning}), single-turn conversation (\textit{LLaVA-Bench-conversation}), and multi-turn dialogue (\textit{MM-MT}). Groc-PO shows superiority.}
  \label{tab:reason}
  \centering
  \small
  \setlength{\tabcolsep}{1.75pt}
  \renewcommand{\arraystretch}{0.95}
  \begin{tabular}{lcccccc}
    \toprule
    Models
      & \shortstack{LLaVA-\\complex\\ reason~\cite{liu2023visual}}
      & \shortstack{LLaVA-\\conversation\\~\cite{liu2023visual}}
      & \shortstack{SEED-\\reason\\~\cite{li2023seed}}
      & \shortstack{MME-\\reason(\%)\\~\cite{fu2023mme}}
      & \shortstack{MM\\-MT\\~\cite{agrawal2024pixtral}} \\
    \midrule
    DPO & 56.9 & 62.0 & 74.3 & 38.3 &  1.88 \\ 
    \rowcolor{cyan!12}
    \textbf{Groc-PO} & \textbf{82.5} & \textbf{67.2} & \textbf{76.5} & \textbf{51.8} & \textbf{2.55} \\ 
     & (+45\%) & (+8\%) & (+3\%) & (+35\%) & (+36\%) \\
    \bottomrule
  \end{tabular}
\end{table}

\subsection{Ablation Study}
\label{sec:ablation}
\subsubsection{\textbf{Contribution of Groc-PO Loss Components.}}
To verify the effectiveness of our Groc-PO Loss, we compared it against 3 variants: (1) \textit{DPO Loss}: uses GCPD data with standard DPO loss; (2) \textit{Stage-aware Only}; and (3) \textit{Difficulty-aware Only}.
Table \ref{tab:loss_ablation} shows that the full Groc-PO Loss achieves the best performance. The individual components each provide significant gains. 

In addition, We also performed ablation studies on the key hyperparameters of the loss, $\alpha$ (Stage-aware) and $\eta$ (Hardness-aware).

\begin{table}[t]
  \caption{\textbf{Ablation study} of the loss components and hyperparameters ($\alpha, \eta$) on MM-Hal~\cite{sun2023aligning}, as described in Section~\ref{sec:grocpo_loss}.}
  \label{tab:loss_ablation}
  \centering
  \small
  \setlength{\tabcolsep}{7pt} 
  \renewcommand{\arraystretch}{0.85} 

  \begin{tabular}{lcccc}
    \toprule
  \multicolumn{1}{c}{\raisebox{0.8ex}[0pt]{\textbf{Loss Setting}}} &
  \raisebox{0.8ex}[0pt]{\textbf{$\alpha$}} &
  \raisebox{0.8ex}[0pt]{\textbf{$\eta$}} &
    \shortstack{MM-Hal\\Score ($\uparrow$)} &
    \shortstack{MM-Hal\\Rate ($\downarrow$)} \\
    \midrule

    \multicolumn{5}{l}{\textit{(A) Component Ablation}} \\

    \hspace{1em}DPO Loss & 0 & 0 & 2.24 & 60.0 \\

    \hspace{1em}+ Stage-aware & 0.25 & 0 & 2.46 & 55.0 \\
    \hspace{1em}+ Difficulty-aware & 0 & 2 & 2.42 & 57.0 \\

    \rowcolor{cyan!12}
    \hspace{1em}+ \textbf{Groc-PO Loss} & \textbf{0.25} & \textbf{2} & \textbf{2.76} & \textbf{47.0} \\

    \midrule 
    
    \multicolumn{5}{l}{\textit{(B) Sensitivity to $\alpha$}} \\

    \hspace{1em}w/ $\alpha=0$ & 0 & 2 & 2.42 & 57.0 \\

    \rowcolor{cyan!12}
    \hspace{1em}\textbf{w/ $\alpha=0.25$} & \textbf{0.25} & \textbf{2} & \textbf{2.76} & \textbf{47.0} \\

    \hspace{1em}w/ $\alpha=0.5$ & 0.5 & 2 & 2.40 & 55.5 \\
    
    \midrule

    \multicolumn{5}{l}{\textit{(C) Sensitivity to $\eta$ }} \\
    \hspace{1em}w/ $\eta=0$ & 0.25 & 0 & 2.46 & 55.0 \\
    \hspace{1em}w/ $\eta=1$ & 0.25 & 1 & 2.57 & 51.0 \\
    \rowcolor{cyan!12}
    \hspace{1em}\textbf{w/ $\eta=2$} & \textbf{0.25} & \textbf{2} & \textbf{2.76} & \textbf{47.0} \\
    \bottomrule
  \end{tabular}
\end{table}

\begin{figure*}[t!]
  \centering
  \includegraphics[width=0.95\linewidth]{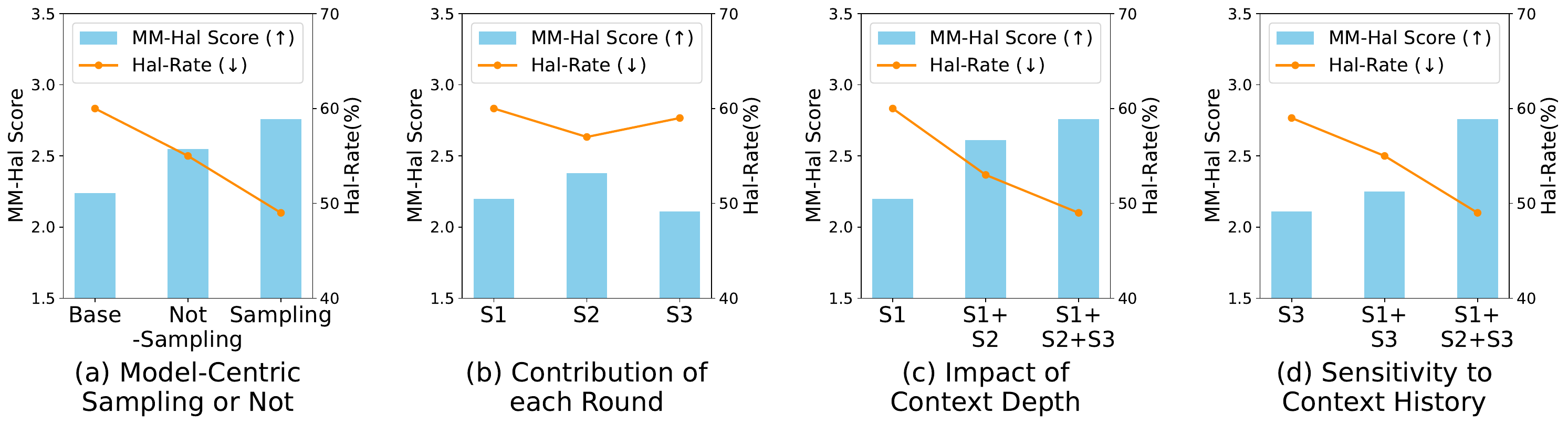}
  \caption{\textbf{Ablations and Analysis.} (a) Effectiveness of Model-Centric Sampling or not. (b) Contribution of individual stages: S2 peaks while S3 degrades from misalignment without history. (c) Impact of context depth: monotonic improvement with progressive history. (d) Sensitivity to history length: performance on a fixed stage-3 query improves as context is added.}
  \label{fig:four_plots}
\end{figure*}

\subsubsection{\textbf{Effect of Model-Centric Sampling.}}
We conduct an ablation of the data construction strategy, comparing a model trained solely on Teacher-generated preference data with one trained on our final dataset using \texttt{"}model-centric sampling\texttt{"}.
Figure~\ref{fig:four_plots}a shows that the latter performs better, suggesting that learning from its own imperfect responses provides a closer data distribution and more targeted alignment signals, thereby improving self-alignment.

\subsubsection{\textbf{Contribution of Each Stage.}}
To quantify the contribution of each context stage, we established three independent training settings: $\text{S1}$, $\text{S2}$, and $\text{S3}$. These models were exclusively trained on data from their respective stages, lacking historical context in $\text{S2}$ and $\text{S3}$. 

Figure~\ref{fig:four_plots}b revealed that $\text{S2}$ outperformed $\text{S1}$ and $\text{S3}$. 
Because $\text{S3}$ relies on preceding information, the absence of context leads to misalignment and performance decline. This affirms the necessity of collaboration among three context rounds, asserting that optimal performance requires structured integration.

\subsubsection{\textbf{Impact of Multi-stage Context Depth.}}
To investigate the influence of contextual learning depth, we used three training settings: (1) using the 1st-round data (S1-Only); (2) using the data of 1st and 2nd rounds (S1+S2); and (3) using full data of 3 stages. 
Figure~\ref{fig:four_plots}c shows the model performance monotonically improves with increasing context depth and complexity.

This demonstrates that our designed \texttt{"}perception$\rightarrow$understanding$\rightarrow$reasoning\texttt{"} progressive learning path is indispensable for building model's compressive capabilities and faithfulness.

\begin{figure}[t]
  \centering
  \includegraphics[width=\linewidth]{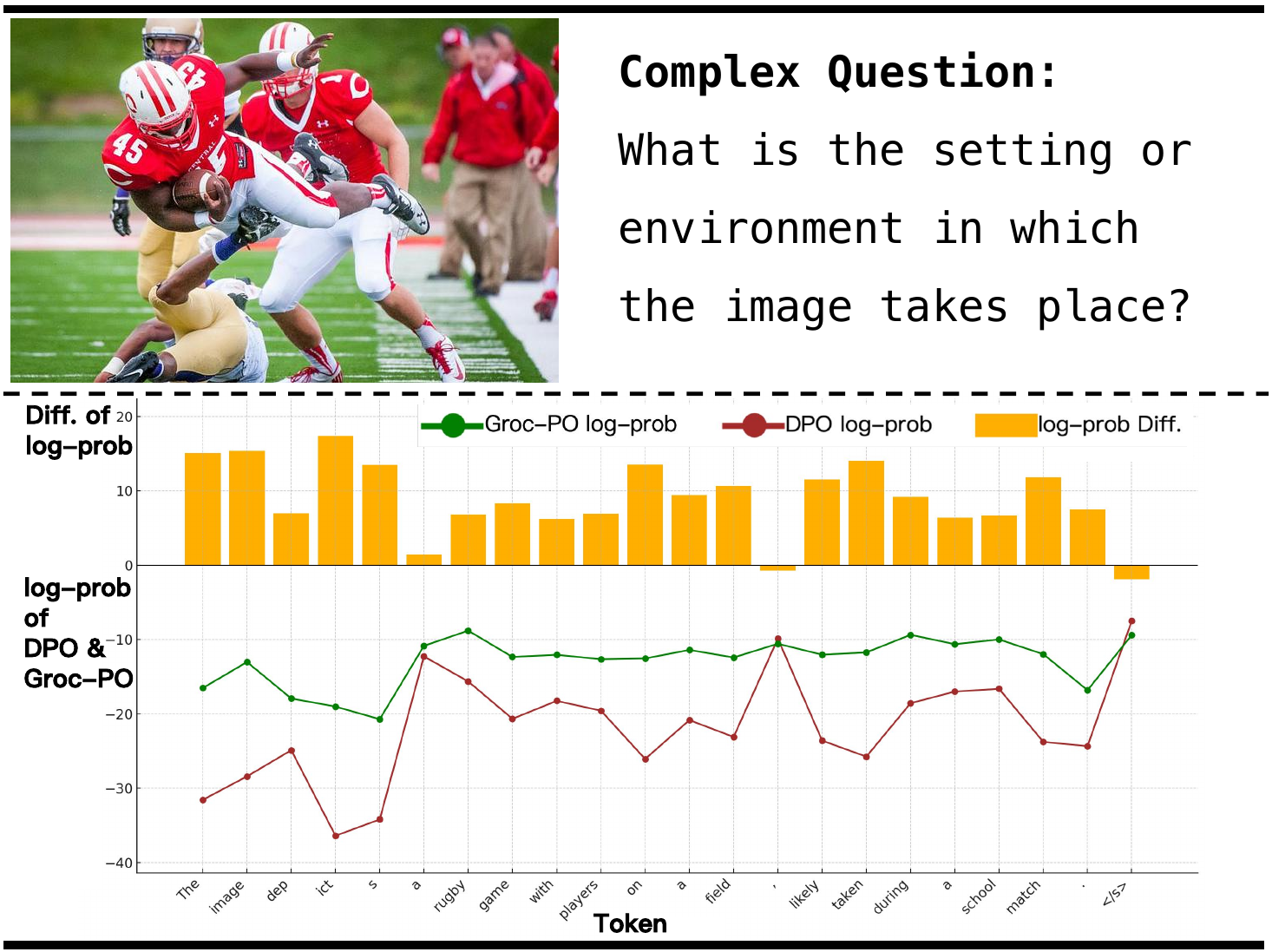}
  \caption{\textbf{Token-level log-probability (log-prob) comparison on chosen responses $y_r^+$ for complex tasks:} Groc-PO vs. DPO. The curves (green:Groc-PO, brown:DPO) show log-prob assigned to each token. The top yellow bar chart illustrates the difference (Groc-PO - DPO), which is almost entirely positive. This suggests Groc-PO exhibits higher internal confidence on complex tasks.}
  \label{fig:log_formal}
\end{figure}

\begin{table}[t]
  \caption{\textbf{Comparison of training with and without grounded context.} The $\texttt{"}$Flattened DPO$\texttt{"}$ (the same 3-stage 17k dataset but without grounded context) shows lower faithfulness than Groc-PO (with grounded context).}
  \label{tab:flatted}
  \centering
  \small
  \setlength{\tabcolsep}{6.5pt} 
  \renewcommand{\arraystretch}{0.85}
  \begin{tabular}{l|cc}
    \toprule
    \textbf{Setting} & \shortstack{MM-Hal Score($\uparrow$)} & \shortstack{Hal-Rate($\downarrow$)} \\
    \midrule
    Flattened DPO & 2.17 & 61.0 \\
    \midrule
    \rowcolor{cyan!12}
    \textbf{Full Structured Context} & \textbf{2.76} & \textbf{47.0} \\
    \bottomrule
  \end{tabular}
\end{table}

\begin{table}[!t]
\caption{\textbf{Training Overhead Comparison.} The average per-sample processing time and peak memory usage in DPO and Groc-PO, indicating that Groc-PO adds only minor overhead.}
\label{tab:overhead}
\centering
\small
\setlength{\tabcolsep}{4pt}
\renewcommand{\arraystretch}{0.85}
\begin{tabular}{@{}l|ccc@{}}
\toprule
\textbf{Model} & Avg. time (s)  & Peak Memory Usage (MB) \\
\midrule
DPO & 1.99  & 40420 \\
\rowcolor{cyan!12}
\textbf{Groc-PO (Ours)} &  \textbf{2.05 (+3\%)} &   \textbf{40432 (+0.03\%)} \\
\bottomrule
\end{tabular}
\end{table}

\begin{figure}[t]
  \centering
  \includegraphics[width=\linewidth]{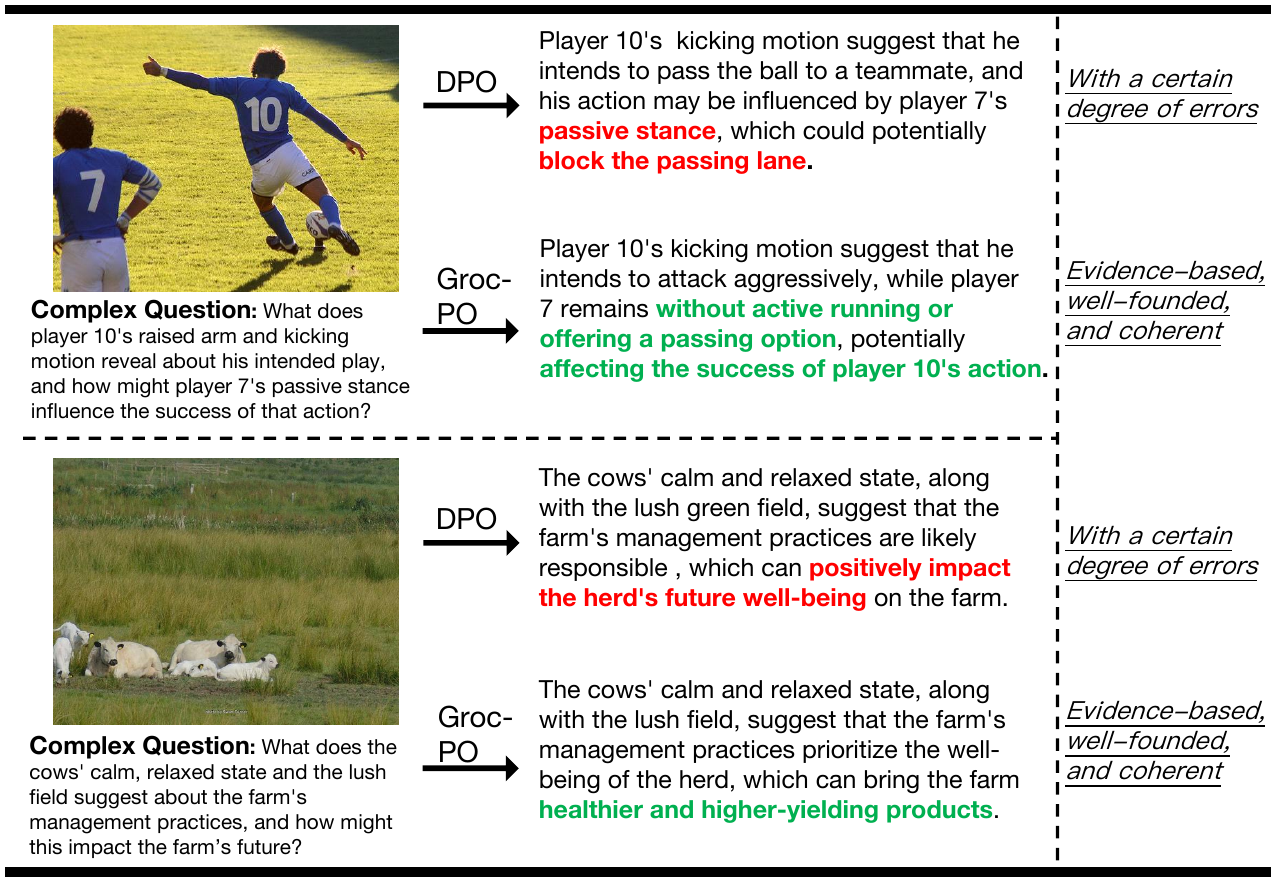}
  \caption{\textbf{Comparison on complex tasks:} Groc-PO vs. DPO. While DPO shows certain degree of errors, Groc-PO demonstrates robust, evidence-based, and coherent reasoning.}
  \label{fig:visual1}
\end{figure}

\subsection{Analysis and Discussion}
\label{sec:analysis}
\subsubsection{\textbf{Dependence on Structured Context.}}
To verify the importance of structured context history, we designed a baseline named \texttt{"}Flattened DPO\texttt{"}, which uses the same 17k preference pairs but removes all context history, degrading all training data to single-stage question-answer pairs. Table \ref{tab:flatted} shows the \texttt{"}Flattened DPO\texttt{"} is far below structured context Groc-PO.
A likely reason is that many Stage 3 require information from the first two stages, and removing history has left the model misaligned.
This result underscores the necessity of incorporating structured context to cultivate contextual coherence and enhance faithfulness.

\subsubsection{\textbf{Sensitivity to History Length.}}
We evaluated the model's ability to leverage contextual history of varying lengths by testing the same third-stage question (S3) under three conditions: 
\textbf{Zero-shot} (image + S3), \textbf{1-stage context} (image + S1 + S3), and \textbf{2-stage context} (full history: image + S1 + S2 + S3). 
Figure~\ref{fig:four_plots}d shows that performance exhibited a clear improvement: $2\text{-stage} > 1\text{-stage} > \text{Zero-shot}$. This empirically validates the critical role of multi-stage context in multimodal dialogue, demonstrating that more complete history enables the model to better localize the question, perform logical reasoning, and generate accurate responses.

\subsubsection{\textbf{Training Overhead.}}
We compared the training overhead of Groc-PO with DPO. Table~\ref{tab:overhead} reports the average processing time per sample and peak memory usage. 
Results show that Groc-PO introduces only marginal overhead.

\subsubsection{\textbf{Discussion: Effectiveness of Grounded Context Supervision.}}
Our analysis highlights the effectiveness of grounded context supervision. As shown in Figure~\ref{fig:four_plots}c, model performance consistently improves as more complete grounded context is incorporated during training. This suggests that more targeted and stage-specific supervision on early grounding stages may improve the reliability of later-stage reasoning. Notably, the improvement is most evident on complex reasoning tasks. As reported in Table~\ref{tab:reason}, the full Groc-PO achieves clear gains on these tasks. At the same time, it maintains strong performance on general and faithfulness benchmarks. These results indicate that grounded context supervision improves reasoning quality, is beneficial for challenging multimodal tasks, and maintains competitive performance on general and faithfulness benchmarks.

\subsection{Case Study and Visualization}
\label{sec:visualization}

\subsubsection{\textbf{Token-level Log-Probabilities on Chosen Responses of Complex Tasks: Groc-PO vs. DPO}}
To further investigate Groc-PO's generation confidence for complex tasks, we conducted a case study for token-level log-probability (log-prob) comparison.

In Figure \ref{fig:log_formal}, we observe that Groc-PO (green curve) assigns a higher log-prob to the majority of tokens in the chosen responses compared to DPO (brown curve), where the difference (yellow bars) is almost entirely positive. This indicates that Groc-PO exhibits higher internal confidence when handling complex tasks. 

\subsubsection{\textbf{Visualization of Qualitative Comparison.}}

Figure \ref{fig:visual1} visually confirms that while the DPO relatively fails on complex queries, Groc-PO generates logically accurate responses that are well-supported by explicit visual evidence.